\documentclass[conference]{IEEEtran}
\IEEEoverridecommandlockouts
\usepackage{cite}
\usepackage{amsmath,amssymb,amsfonts}
\usepackage{algorithmic}
\usepackage{graphicx}
\usepackage{textcomp}
\usepackage{xcolor}
\usepackage{multirow}
\usepackage{url}
\usepackage{hyperref}
\ifCLASSOPTIONcompsoc
\usepackage[caption=false,font=normalsize,labelfon
t=sf,textfont=sf]{subfig}
\else
\usepackage[caption=false,font=footnotesize]{subfig}
\fi

\def\BibTeX{{\rm B\kern-.05em{\sc i\kern-.025em b}\kern-.08em
    T\kern-.1667em\lower.7ex\hbox{E}\kern-.125emX}}
\begin{document}

\title{Capturing Multivariate Dependencies of EV Charging Events: From Parametric Copulas to Neural Density Estimation\\
\thanks{This work was partially supported by the following projects:
V4Grid - The Visegrad group for Vehicle to X: Interreg Central Europe Programme,
project No. CE0200803;  LorAI - Low Resource Artificial Intelligence: Horizon Europe Programme, GA No. 101136646.}
}

\author{\IEEEauthorblockN{Martin V\'yboh}
\IEEEauthorblockA{\textit{Kempelen Institute of Intelligent Technologies} \\
Bratislava, Slovakia \\
martin.vyboh@kinit.sk}
\and
\IEEEauthorblockN{Gabriela Grmanov\'a}
\IEEEauthorblockA{\textit{Kempelen Institute of Intelligent Technologies} \\
Bratislava, Slovakia \\
gabriela.grmanova@kinit.sk}
}

\maketitle

\begin{abstract}
Accurate event-based modeling of electric vehicle (EV) charging is essential for grid reliability and smart-charging design. While traditional statistical methods capture marginal distributions, they often fail to model the complex, non-linear dependencies between charging variables, specifically arrival times, durations, and energy demand. This paper addresses this gap by introducing the first application of Vine copulas and Copula Density Neural Estimation framework (CODINE) to the EV domain. We evaluate these high-capacity dependence models across three diverse real-world datasets. Our results demonstrate that by explicitly focusing on modeling the joint dependence structure, Vine copulas and CODINE outperform established parametric families and remain highly competitive against state-of-the-art benchmarks like conditional Gaussian Mixture Model Networks. We show that these methods offer superior preservation of tail behaviors and correlation structures, providing a robust framework for synthetic charging event generation in varied infrastructure contexts.
\end{abstract}


\section{Introduction}
The adoption of electric vehicles (EVs) is steadily increasing, with the EU stock reaching approximately 5.87 million by the end of 2024 \cite{Eurostat2025}. The EV30@30 initiative targets a 30\% share of EVs in new car sales by 2030 \cite{IEA}. The growing penetration of EVs presents both challenges, such as increased energy demand, and opportunities for grid management through smart charging and vehicle-to-grid strategies, which can shift charging to off-peak periods and provide flexibility to support the power system \cite{Mishra2025Smart}. This expanded role increases the need for accurate event-based models that capture individual vehicle dynamics (arrival, charging, departure), enabling the design and evaluation of such strategies and their impact on load management and grid reliability \cite{Lin2025Electric}.

\subsection{Related Work}
Approaches for modeling the EV charging events can be categorized into three types: \emph{statistical characterization}, \emph{stochastic processes} and \emph{Machine learning (ML) methods} \cite{AmaraOuali2021Review}, an organization similar to those considered in studies \cite{Lin2025Electric, Zhao2026EVBehavior}.

\emph{Statistical methods} model EV charging events using probabilistic distributions, such as kernel density estimation (KDE) for arrival and staying times and charging capacity \cite{Chen2020Modeling}, or mixture models (e.g., Gaussian mixtures) for connection time and required energy \cite{Lahariya2020Synthetic}.
While these approaches capture marginal distributions well, they fail to fully represent dependencies between variables (e.g., late arrival vs. shorter charging duration). Copula-based methods \cite{Sklar1959Copulas} address this limitation by separating the dependence structure from the marginals, allowing the joint behavior of variables to be modeled independently of their individual distributions. In the context of EV charging event modeling, Chen et al. applied two-dimensional copulas to model correlations of charging start time and duration, and charging duration and charged capacity \cite{Chen2018Analysis}. Study \cite{Chen2020Modeling} compares ternary KDE and joint method based on edge KDE and ternary Student-t copula function, but the evaluation is based solely on in-sample performance without assessing generalization. Einolander and Lahdelma evaluated five common copula families \cite{Einolander2022Copula}, finding that elliptical copulas, such as Gaussian and Student‑t, outperformed simpler Archimedean copulas like Clayton, Frank, or Gumbel. More flexible models exist, such as Vine copulas, which decompose multivariate dependencies into products of bivariate copulas arranged in a nested tree structure \cite{Aas2009PairCopula}. This allows different copula families to be used for individual pairwise relationships, thereby enabling a detailed representation of heterogeneous dependencies. Such flexibility could benefit event-level EV charging simulations.

\emph{Stochastic processes} \cite{AmaraOuali2021Review} generate events by sampling from probabilistic distributions of key parameters. 

\emph{ML methods} include regression-based learners, tree ensembles, and deep learning architectures such as recurrent neural networks and LSTMs, which are well-suited for capturing sequential and contextual patterns in charging data \cite{AmaraOuali2021Review}. More recent studies have also explored transformer-based and generative AI models for this task \cite{Zhao2026EVBehavior}. 

The literature includes approaches that integrate statistical modeling and ML to capture complex dependencies \cite{Fahradi2023ReviewEV}, e.g., Li et al. \cite{Li2022SynthesisEV} train conditional density networks to model and generate synthetic events. However, these methods often rely on a chain of 1D estimations rather than a unified joint density. Zeng and Wang proposed a neural network framework for high-dimensional copula estimation, but it requires training multiple networks (growing with the number of features) and constraint-based training to ensure a statistically valid copula representation \cite{Zeng2022NeuralCopula}. Letizia et al. introduced the Copula Density Neural Estimation (CODINE) method \cite{Letizia2025Codine}, a neural network-based approach that directly learns the copula density using normalizing transformations and neural density estimation. The method models complex and high-dimensional dependence structures while ensuring that the resulting function satisfies the properties of a valid copula. By combining neural network flexibility with the theoretical framework of copulas, CODINE enables scalable estimation of dependence structures without requiring specification of parametric copula families.

\subsection{Our Contribution}
We contribute to the field by simultaneously introducing a unified joint-dependency framework for EV events using Vine copulas and CODINE, and validating this approach across three diverse charging datasets: home, private garage, and public infrastructure. We demonstrate that these joint-modeling approaches better preserve tail behavior and correlation structures than traditional parametric copulas and the established GMMNet benchmark \cite{Li2022SynthesisEV}.

\begin{table*}[t]
\caption{Performance Comparison Across Datasets (Mean $\pm$ SD over 5 Random Seeds). Joint-dependency Methods are in Bold; the Best-Performing Model per Dataset (Lowest Avg. Rank) is Underlined.}
\centering
\small
\setlength{\tabcolsep}{3pt}
\begin{tabular*}{\textwidth}{@{\extracolsep{\fill}}|c|c||c|c|c|c|c|c|c||c|}
\hline
\textbf{DS} & \textbf{Model} & \textbf{NLL$^{\mathrm{a}}$} & \textbf{$\tau\text{-Diff}$} & \textbf{$\rho_1$} & \textbf{$\rho_2$} & \textbf{$\text{MAE}_{\text{LT}}$} & \textbf{$\text{MAE}_{\text{UT}}$} & \textbf{$\text{MAE}_{\text{Load}}$} & \textbf{Avg. Rank} \\
\hline
\multirow{8}{*}{\rotatebox[origin=c]{90}{Proprietary}} 
& Clayton   & $9.54$            & $0.49 \pm 0.08$         & $1.68 \pm 0.32$ & $1.12 \pm 0.09$ & $0.36 \pm 0.04$ & $0.05 \pm 0.06$ & $1.26 \pm 0.22$ & $5.00$ \\
& \textbf{CODINE}    & $\mathbf{10.89}$  & $\mathbf{0.30 \pm 0.16}$ & $\mathbf{1.89 \pm 0.41}$ & $\mathbf{1.29 \pm 0.09}$ & $\mathbf{0.03 \pm 0.06}$ & $\mathbf{0.09 \pm 0.06}$ & $\mathbf{0.68 \pm 0.12}$ & $\mathbf{4.57}$ \\
& Frank     & $10.77$           & $0.46 \pm 0.10$         & $2.73 \pm 0.10$ & $0.84 \pm 0.07$ & $0.21 \pm 0.06$ & $0.09 \pm 0.09$ & $1.22 \pm 0.15$ & $5.57$ \\
& Gaussian  & $9.10$            & $0.25 \pm 0.08$         & $1.63 \pm 0.14$ & $1.16 \pm 0.07$ & $0.11 \pm 0.11$ & $0.17 \pm 0.12$ & $1.02 \pm 0.08$ & $3.43$ \\
& GMMNet    & $15.29$           & $0.18 \pm 0.07$         & $1.64 \pm 0.27$ & $1.40 \pm 0.05$ & $0.14 \pm 0.05$ & $0.07 \pm 0.05$ & $1.11 \pm 0.40$ & $4.57$ \\
& Gumbel    & $9.33$            & $0.50 \pm 0.06$         & $1.64 \pm 0.22$ & $1.06 \pm 0.05$ & $0.17 \pm 0.08$ & $0.35 \pm 0.22$ & $1.22 \pm 0.26$ & $5.00$ \\
& Student-t & $9.19$            & $0.27 \pm 0.07$         & $1.46 \pm 0.18$ & $1.15 \pm 0.03$ & $0.35 \pm 0.03$ & $0.33 \pm 0.15$ & $2.13 \pm 1.17$ & $4.57$ \\
& \underline{\textbf{Vine}} & \underline{$\mathbf{9.58}$} & \underline{$\mathbf{0.21 \pm 0.03}$} & \underline{$\mathbf{1.61 \pm 0.25}$} & \underline{$\mathbf{1.30 \pm 0.06}$} & \underline{$\mathbf{0.07 \pm 0.05}$} & \underline{$\mathbf{0.09 \pm 0.06}$} & \underline{$\mathbf{0.89 \pm 0.13}$} & \underline{$\mathbf{3.29}$} \\
\hline
\hline
\multirow{8}{*}{\rotatebox[origin=c]{90}{Trondheim}} 
& Clayton   & $9.84$            & $0.33 \pm 0.04$         & $1.58 \pm 0.21$ & $0.96 \pm 0.02$ & $0.19 \pm 0.05$ & $0.06 \pm 0.02$ & $0.37 \pm 0.05$ & $4.29$ \\
& \textbf{CODINE}    & $\mathbf{9.90}$   & $\mathbf{0.19 \pm 0.04}$ & $\mathbf{1.40 \pm 0.37}$ & $\mathbf{1.10 \pm 0.02}$ & $\mathbf{0.16 \pm 0.05}$ & $\mathbf{0.03 \pm 0.01}$ & $\mathbf{0.32 \pm 0.19}$ & $\mathbf{3.29}$ \\
& Frank     & $10.92$           & $0.32 \pm 0.04$         & $2.05 \pm 0.16$ & $0.84 \pm 0.01$ & $0.25 \pm 0.02$ & $0.06 \pm 0.03$ & $0.46 \pm 0.14$ & $6.86$ \\
& Gaussian  & $10.12$           & $0.15 \pm 0.05$         & $1.72 \pm 0.32$ & $0.96 \pm 0.02$ & $0.21 \pm 0.02$ & $0.10 \pm 0.03$ & $0.43 \pm 0.05$ & $5.14$ \\
& \underline{GMMNet} & \underline{$9.67$} & \underline{$0.12 \pm 0.06$} & \underline{$1.47 \pm 0.25$} & \underline{$1.29 \pm 0.03$} & \underline{$0.11 \pm 0.02$} & \underline{$0.02 \pm 0.02$} & \underline{$0.27 \pm 0.11$} & \underline{$2.29$} \\
& Gumbel    & $10.19$           & $0.33 \pm 0.05$         & $1.59 \pm 0.20$ & $0.94 \pm 0.02$ & $0.26 \pm 0.03$ & $0.19 \pm 0.07$ & $0.39 \pm 0.06$ & $6.14$ \\
& Student-t & $10.18$           & $0.16 \pm 0.05$         & $1.53 \pm 0.16$ & $0.85 \pm 0.01$ & $0.16 \pm 0.03$ & $0.31 \pm 0.09$ & $1.16 \pm 0.55$ & $5.57$ \\
& \textbf{Vine}      & $\mathbf{9.84}$   & $\mathbf{0.15 \pm 0.04}$ & $\mathbf{1.62 \pm 0.22}$ & $\mathbf{0.99 \pm 0.03}$ & $\mathbf{0.08 \pm 0.04}$ & $\mathbf{0.04 \pm 0.02}$ & $\mathbf{0.28 \pm 0.04}$ & $\mathbf{2.43}$ \\
\hline
\hline
\multirow{8}{*}{\rotatebox[origin=c]{90}{Dundee}} 
& Clayton   & $8.70$            & $0.72 \pm 0.00$         & $1.41 \pm 0.04$ & $0.87 \pm 0.01$ & $0.40 \pm 0.01$ & $0.14 \pm 0.01$ & $12.30 \pm 0.06$ & $5.00$ \\
& \underline{\textbf{CODINE}} & \underline{$\mathbf{8.17}$} & \underline{$\mathbf{0.09 \pm 0.02}$} & \underline{$\mathbf{1.59 \pm 0.33}$} & \underline{$\mathbf{1.18 \pm 0.01}$} & \underline{$\mathbf{0.06 \pm 0.04}$} & \underline{$\mathbf{0.06 \pm 0.00}$} & \underline{$\mathbf{2.13 \pm 0.09}$} & \underline{$\mathbf{3.00}$} \\
& Frank     & $17.44$           & $0.88 \pm 0.01$         & $3.46 \pm 0.02$ & $0.74 \pm 0.01$ & $0.27 \pm 0.00$ & $0.13 \pm 0.01$ & $13.45 \pm 0.16$ & $7.29$ \\
& Gaussian  & $8.83$            & $0.07 \pm 0.01$         & $1.42 \pm 0.04$ & $0.92 \pm 0.01$ & $0.10 \pm 0.00$ & $0.07 \pm 0.01$ & $12.46 \pm 0.04$ & $3.14$ \\
& GMMNet    & $8.36$            & $0.12 \pm 0.02$         & $2.03 \pm 0.13$ & $1.16 \pm 0.05$ & $0.19 \pm 0.01$ & $0.12 \pm 0.01$ & $4.25 \pm 0.31$ & $4.29$ \\
& Gumbel    & $10.15$           & $0.73 \pm 0.00$         & $1.39 \pm 0.03$ & $0.86 \pm 0.00$ & $0.29 \pm 0.01$ & $0.21 \pm 0.02$ & $12.13 \pm 0.04$ & $5.57$ \\
& Student-t & $11.33$           & $0.09 \pm 0.01$         & $1.38 \pm 0.03$ & $0.91 \pm 0.01$ & $0.20 \pm 0.00$ & $0.18 \pm 0.01$ & $9.46 \pm 0.14$ & $4.29$ \\
& \textbf{Vine}      & $\mathbf{9.41}$   & $\mathbf{0.07 \pm 0.01}$ & $\mathbf{1.44 \pm 0.02}$ & $\mathbf{0.95 \pm 0.01}$ & $\mathbf{0.04 \pm 0.00}$ & $\mathbf{0.09 \pm 0.01}$ & $\mathbf{12.70 \pm 0.11}$ & $\mathbf{3.43}$ \\
\hline
\multicolumn{10}{l}{$^{\mathrm{a}}$NLL calculated via KDE fitted on generated and evaluated on test samples. See Section \ref{sec:evaluation_framework} for details.}
\end{tabular*}
\label{tab_results}
\end{table*}

\section{Data and Experimental Setup}
We evaluate our models across three EV charging datasets located in Slovakia, Norway (Trondheim), and Scotland (Dundee). Each record consists of three continuous variables: arrival time (decimalized to hours and shifted to capture overnight charging continuity), charging duration (decimal hours), and energy consumed (kWh). Shifting the origin of the arrival time to 6 AM ensures overnight sessions do not suffer from the overnight circularity discontinuity.

For all experiments, we employ a chronological temporal split: 80\% for training, 10\% for validation, and 10\% for testing. To maintain computational efficiency during the evaluation of the expansive Dundee dataset, we utilize a common random subsample of $20\,000$ records for metric calculation, while models are trained on the full available history.

\indent \textit{1) Proprietary Dataset (Residential):} Comprises $1\,097$ sessions from a residential AC charger in Slovakia, collected between December 2022 and October 2024. It represents a small-scale environment characterized by distinct day/night patterns. Here, the charging station often doubles as a long-term parking space. Consequently, the connection duration exhibits significant variance.

\indent \textit{2) Trondheim Dataset (Multi-User Residential):} Initially includes a mix of public and private EV charging sessions across 24 Norwegian residential garages, recorded from December 2018 to January 2020 \cite{Sorensen2021Trondheim}. To keep a focus on private charging behavior, we filtered the data to include only privately owned EV charging points. Selection resulted in $5\,442$ unique sessions. This allows us to evaluate the models' ability to estimate the behavior of multiple private users.

\indent \textit{3) Dundee Dataset (Public):} Consists of public charging records published by the City of Dundee \cite{Dundee2024Dundee}. We extracted sessions from January 2023 to December 2024, keeping only chargers with the flag ``rapid'' to isolate charging events with different power. This yielded a dataset of $166\,888$ records. Incorporating this data allows for experiments on a distribution that differs from residential datasets both in size and character.

\section{Methodology}
This study presents a comprehensive evaluation framework for modeling multivariate dependencies in EV charging datasets through copula-based generative models. Our methodology encompasses three distinct modeling paradigms: (1) neural copula density estimation via adversarial training (CODINE), (2) neural Gaussian Mixture Model networks with feature-specific architectures (GMMNet), and (3) parametric copula families with Vine copulas. We model three previously introduced variables from EV charging session records: arrival time, charging duration, and energy consumed. 

The experimental framework follows a systematic approach: model training with an early stopping mechanism, sample generation, and comprehensive evaluation across multiple performance metrics. We assess models on distributional similarity, dependence structure/tail behavior modeling, generalization, and practical load forecasting accuracy. We ensure robustness is taken into account through multi-seed runs.

\subsection{Model Architectures}
\subsubsection{CODINE}
Utilizes a discriminator network composed of three 100-unit hidden layers and LeakyReLU activations \cite{Letizia2025Codine}. We employ the GAN divergence from this study, as our preliminary tests showed it is most stable for this domain. The discriminator learns the copula density by distinguishing empirical pseudo-observations from uniform noise, effectively learning the copula density. Optimization uses Adam with a batch size of 64.

\subsubsection{GMMNet}
Following architectural optimizations identified in \cite{Li2022SynthesisEV}, we utilize an ensemble of $d$ conditional mixture density networks. Each comprises two 32-unit hidden layers with batch normalization and ReLU activations. The output layer predicts parameters for a 5-component GMM. We use Adam (batch 64) to minimize the Negative Log-likelihood (NLL). To address numerical instabilities encountered during the replication study, we applied gradient and output clamping to ensure model stability.

\subsubsection{Parametric Copulas}
We implement six reference families: Gaussian, Clayton, Frank, Gumbel, Student-t, and R-Vine. Archimedean and elliptical copulas' parameters are estimated via the inverse Kendall’s Tau method. For R-Vines, we utilize a Regular Vine structure where the $d$-dimensional density is decomposed into $d(d-1)/2$ bivariate pair-copulas. Structure selection and parameter estimation are performed sequentially using maximum likelihood to determine optimal pair-copula families at each tree level.

\subsection{Training Procedure}
We implement early stopping for CODINE and GMMNet based on $\tau\text{-Diff}$ with a 500-epoch patience to prevent overfitting. To ensure statistical robustness, all experiments are conducted across 5 unique random seeds. The best-performing seed (based on the lowest $\tau\text{-Diff}$) is selected for qualitative evaluation and the generation of plots.

\begin{figure*}[t]
    \centering
    \subfloat[Real data\label{fig:trond_real}]{%
        \includegraphics[width=0.32\textwidth]{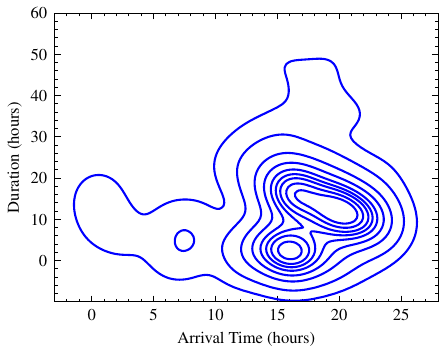}}
    \hfill
    \subfloat[Clayton copula\label{fig:trond_clayton}]{%
        \includegraphics[width=0.32\textwidth]{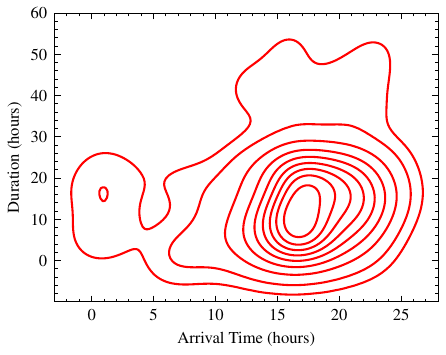}}
    \hfill
    \subfloat[CODINE\label{fig:trond_codine}]{%
        \includegraphics[width=0.32\textwidth]{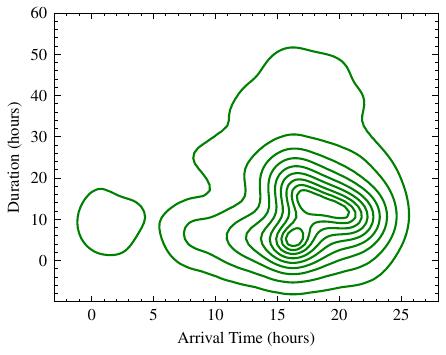}}
    \caption{Bivariate density contours for Arrival Time vs. Duration (Trondheim dataset). The data exhibits a multimodal distribution that the parametric copula fails to capture, whereas the neural-based CODINE recovers the multimodal dependency structure.}
    \label{fig:trondheim_multimodal}
\end{figure*}

\subsection{Evaluation Framework} \label{sec:evaluation_framework}
We evaluate model performance across four aspects: likelihood, dependence preservation, generalization, and grid impact. For each model, we generate a synthetic dataset equal in size to the test set. Following \cite{Li2022SynthesisEV, Letizia2025Codine}, data generation for CODINE and GMMNet is performed via Gibbs sampling.

\subsubsection{Density and Distributional Fidelity} The primary metric is the NLL. For a test set $\{ \mathbf{x}_k \}_{k=1}^n$ with joint density $f(\mathbf{x})$, $\text{NLL} = -\frac{1}{n} \sum_{k=1}^{n} \log f(\mathbf{x}_k)$. Since GMMNet estimates conditional densities, joint densities are approximated using KDE (Gaussian kernel, optimal bandwidth found via 5-fold cross-validation) fitted on generated samples and evaluated on the test set. 

To quantify univariate distributional similarity, we compute the Average Wasserstein Distance ($\rho_1$) as defined in \cite{Li2022SynthesisEV}:
\begin{displaymath}
\rho_1 = \frac{1}{d} \sum_{i=1}^{d} \int_{-\infty}^{\infty} |F_i(x) - \hat{F}_i(x)| \,dx,
\end{displaymath}
where $F_i$ and $\hat{F}_i$ represent the empirical cumulative distribution functions of the real and synthetic data, for the $i$-th feature, respectively.

\subsubsection{Dependence Structure and Tail Behavior}
To evaluate the preservation of the underlying dependence structure, we compare the rank correlation matrices. Let $\mathbf{R}$ and $\hat{\mathbf{R}}$ be the $d \times d$ Kendall's $\tau$ correlation matrices for the real and generated data. The correlation error is quantified using the Frobenius norm of their difference $\tau\text{-Diff} = \|\mathbf{R} - \hat{\mathbf{R}}\|_F$. This metric provides a single scalar value representing the total discrepancy in the pairwise rank-dependence structures across all EV charging session variables. 

To assess the model's ability to replicate extreme co-movements, we evaluate the empirical tail dependence coefficients at quantile threshold $\alpha = 0.95$. For uniform pseudo-observations of the $i$-th and $j$-th features ($U_i$, $U_j$), the upper and lower tail dependence coefficients are defined as:
\begin{displaymath}
\lambda_u = P(U_i > \alpha \mid U_j > \alpha), \mkern8mu \lambda_l = P(U_i < 1-\alpha \mid U_j < 1-\alpha).
\end{displaymath}
The accuracy in capturing these extremal dependencies is quantified using the Mean Absolute Error (MAE) between the empirical coefficients of the real and synthetic data across all pairs, denoted as $\text{MAE}_{\text{UT}}$ and $\text{MAE}_{\text{LT}}$. 

\subsubsection{Generalization} 
We utilize the generalization metric $\rho_2$ to detect potential training data memorization \cite{Li2022SynthesisEV}. For each training sample $\mathbf{t}_k \in T$, we compute the distance to its nearest neighbor in the test set $V$ and the synthetic set $S$:
\begin{displaymath}
d(\mathbf{t}_k, V) = \min_{\mathbf{v} \in V} \|\mathbf{t}_k - \mathbf{v}\|_2, \quad d(\mathbf{t}_k, S) = \min_{\mathbf{s} \in S} \|\mathbf{t}_k - \mathbf{s}\|_2.
\end{displaymath}
The metric $\rho_2$ is defined as the mean ratio:
\begin{displaymath}
\rho_2 = \frac{1}{|T|} \sum_{k=1}^{|T|} \frac{d(\mathbf{t}_k, V)}{d(\mathbf{t}_k, S) + \epsilon},
\end{displaymath}
where $\epsilon > 0$ is a small constant for numerical stability. Values significantly exceeding 1 suggest overfitting.

\subsubsection{Grid Impact}
To evaluate the practical utility of the models for power system analysis, we define $\text{MAE}_{\text{Load}}$. This measures the mean absolute error between the average daily load profiles derived from real and synthetic datasets at one-minute resolution, assuming constant average power. This metric quantifies the model's fidelity in capturing the temporal aggregation of charging events.

\begin{figure*}[t]
    \centering
    \subfloat[Real data\label{fig:real_pseudo}]{%
        \includegraphics[width=0.32\textwidth]{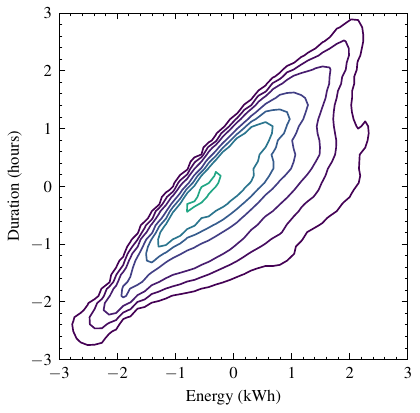}}
    \hfill
    \subfloat[GMMNet\label{fig:gmm_pseudo}]{%
        \includegraphics[width=0.32\textwidth]{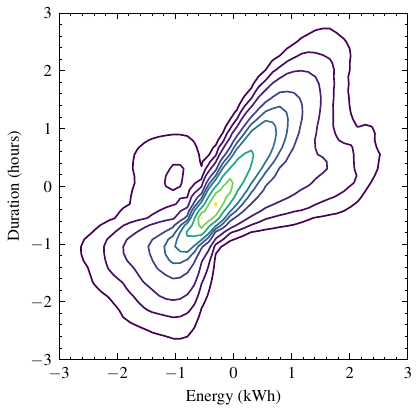}}
    \hfill
    \subfloat[CODINE\label{fig:codine_pseudo}]{%
        \includegraphics[width=0.32\textwidth]{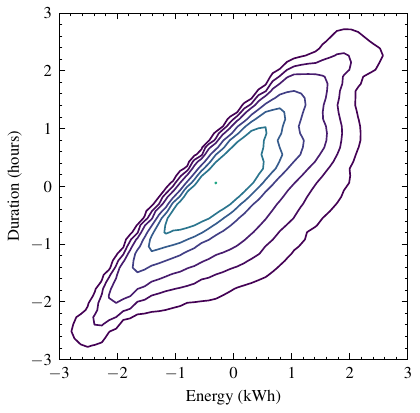}}
    \caption{Comparison of bivariate density contours (Duration vs. Energy) for the Dundee dataset. The plots visualize the z-space, highlighting CODINE's superior ability to capture non-linear dependence and tail structures compared to the GMMNet baseline.}
    \label{fig:copula_comparison}
\end{figure*}

\section{Results}
The performance of the models across the three datasets is summarized in Table \ref{tab_results}, following the metrics in Section \ref{sec:evaluation_framework}.

The results demonstrate that joint-dependency models (Vine copula, CODINE) consistently match or outperform traditional copulas and more effectively capture complex dependency structures. In the Proprietary dataset, the Vine copula achieves the lowest average rank. This suggests that for smaller datasets, Vine copulas offer a more robust estimation of joint density than the neural architectures, which may require more data to converge. While the GMMNet baseline shows strong performance on the Trondheim dataset, the joint-modeling approaches provide more robust results across diverse charging contexts. 

Our joint-dependency methods prove superior in capturing complex multivariate dependencies. Vine copulas and CODINE consistently yield lower $\tau\text{-Diff}$ values than Archimedean families (Clayton, Frank, Gumbel). This indicates that the tree structure of Vine copulas and the neural density estimation methods better replicate the underlying rank-correlation matrices (see Figs. \ref{fig:trondheim_multimodal}, \ref{fig:copula_comparison}). Furthermore, the $\text{MAE}_{\text{LT}}$ and $\text{MAE}_{\text{UT}}$ metrics reveal that neural and hierarchical models are significantly more adept at capturing extreme events, a crucial property for authentically simulating EV charging events. For instance, in the Dundee dataset, CODINE and Vine copulas achieve near-zero tail errors, whereas parametric models fail to capture these extremal co-movements on one or both tails of the distribution.

\begin{figure}[htbp]
    \centering
    \includegraphics[width=\columnwidth]{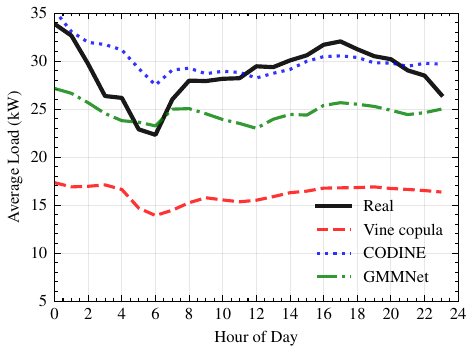}
    \caption{Average daily load curve for the Dundee dataset. While the  Vine underestimates the load magnitude, the neural-based methods exhibit higher fidelity in replicating the average load, with CODINE having lower $\text{MAE}_{\text{Load}}$.}
    \label{fig:load_curves_dundee}
\end{figure}

The $\text{MAE}_{\text{Load}}$ metric serves as a key indicator of the models' practical applicability in grid simulations. On the small-scale Proprietary dataset, CODINE achieves the lowest error ($0.68$), significantly outperforming GMMNet ($1.11$) and Gaussian copula ($1.02$) baselines. While all models produced comparable average load estimates for the Trondheim dataset, CODINE dramatically outperforms all other models on the large-scale Dundee dataset with a $\text{MAE}_{\text{Load}}$ of $2.13$, nearly half that of the GMMNet baseline ($4.25$). This further confirms that capturing the joint density of EV charging events is critical for high-fidelity load estimation (see Figs. \ref{fig:copula_comparison} and \ref{fig:load_curves_dundee}).

Finally, the $\rho_2$ metric remains close to $1.00$ for the methods across all datasets, confirming that the models can somewhat generalize effectively without memorizing the training distribution. In the Dundee dataset, CODINE achieves the lowest NLL, suggesting that the method scales more efficiently to large-scale datasets characterized by intricate dependencies.

\section{Conclusion and Future Work}
This paper presented a comparative evaluation of joint-dependency models for EV charging events. By moving beyond traditional simple parametric copulas, we demonstrated that Vine copulas and CODINE provide a more robust framework for capturing the non-linear, multimodal, and tail-heavy dependencies present in diverse charging infrastructures. Our results across three real-world datasets reveal that these approaches enhance generative fidelity. Notably, the neural-based CODINE model reduced the $\text{MAE}_{\text{Load}}$ by nearly 50\% on the large-scale Dundee dataset compared to state-of-the-art benchmarks. Furthermore, the hierarchical structure of Vine copulas proved most stable in low-data regimes. These findings underscore that capturing the multivariate nature of this task is a prerequisite for accurate synthetic data generation.

Future work may focus on integrating these frameworks into agent-based simulations to evaluate the performance of vehicle-to-grid control algorithms under realistic scenarios. Additionally, we aim to investigate the temporal evolution of EV charging features as they can change over longer horizons.

%
%
%
\bibliographystyle{IEEEtran}
\bibliography{citations}

\end{document}